\documentclass[journal,twoside,web]{ieeecolor}
\usepackage{generic}
\usepackage{cite}
\usepackage{amsmath,amssymb,amsfonts}
\usepackage{algorithmic}
\usepackage{graphicx}
\usepackage{subcaption}
\usepackage{textcomp}
\usepackage[hidelinks]{hyperref}
\usepackage{soul}
\usepackage{float}
\usepackage{adjustbox}
\usepackage{svg}

\usepackage[nonumberlist]{glossaries}
\makeglossaries 
\newacronym{mi}{MI}{Department of Medical Informatics}
\newacronym{umg}{UMG}{University Medical Center Göttingen}
\newacronym{mtb}{MTB}{molecular tumor board}
\newacronym{khentgg}{KHEntgG}{German Hospital Reimbursement Act (\textit{Gesetz über die Entgelte für voll- und teilstationäre Krankenhausleistungen})}
\newacronym{icd}{ICD}{International Statistical Classification of Diseases and Related Health Problems}
\newacronym{ic}{IC}{Information Content}
\newacronym{cs}{CS}{Concept Similarity}
\newacronym{setsim}{SetSim}{Set Similarity}
\newacronym{ehr}{EHR}{electronic health record}

\def\BibTeX{{\rm B\kern-.05em{\sc i\kern-.025em b}\kern-.08em
    T\kern-.1667em\lower.7ex\hbox{E}\kern-.125emX}}
\begin{document}
\bstctlcite{IEEEexample:BSTcontrol}
\title{Improving ICD-based semantic similarity by accounting for varying degrees of comorbidity}
\author{Jan J. Schneider, B. Eng., Marius Adler, MD, Christoph Ammer-Herrmenau, MD, Alexander O. König, MD, Ulrich Sax, PhD and Jonas Hügel, M. Sc.
\thanks{
This research was conducted under the ethics vote  8/1/17 (MolPaC), 11/5/17 (CRU 5002) and 27/8/20 (MTB) from the ethics commission of the University Medical Center Göttingen.
This work was partially funded by the VolkswagenStiftung within the MTB-Report project (Grant: ZN3424), as well by the German Research Foundation (DFG) within the CRU 5002 (Grant: 426671079).
J.Hügel's work was partially funded  by a fellowship within the IFI programme of the German Academic Exchange Service (DAAD) and by the Federal Ministry of Education and Research (BMBF).}
\thanks{J. J. Schneider is with the Department of Medical Informatics, University Medical Center Göttingen,Von-Siebold-Str.3, 37099 Göttingen, Germany (e-mail: janjanosch.schneider@med.uni-goettingen.de).}
\thanks{M. Adler was with University Medical Center Göttingen, Clinic for Gastroenterology, Gastrointestinal Oncology and Endocrinology, Robert-Koch-Str.40, 37075 Göttingen, Germany and is now with Universitätsklinikum Augsburg, III. Medizinische Klinik, Stenglinstr. 2, 86156 Augsburg, Germany (e-mail: marius.adler@uk-augsburg.de).}
\thanks{A. O. König is with the University Medical Center Göttingen, Clinic for Gastroenterology, Gastrointestinal Oncology and Endocrinology, Robert-Koch-Str.40, 37075 Göttingen, Germany (email: alexander.koenig@med.uni-goettingen.de).}
\thanks{C. Ammer-Herrmenau is with the University Medical Center Göttingen, Clinic for Gastroenterology, Gastrointestinal Oncology and Endocrinology, Robert-Koch-Str.40, 37075 Göttingen, Germany (email:christoph.herrmenau@med.uni-goettingen.de). }
\thanks{U. Sax is  with the Department of Medical Informatics, University Medical Center Göttingen, Von-Siebold-Str. 3, 37099 Göttingen, Germany and the Campus Institute Data Science, Georg-August-Universität Göttingen, 37072 Göttingen, Germany (e-mail: ulrich.sax@med.uni-goettingen.de).}
\thanks{J. Hügel is with the Department of Medical Informatics, University Medical Center Göttingen, Von-Siebold-Str. 3, 37099 Göttingen, Germany and the Campus Institute Data Science, Georg-August-Universität Göttingen, 37072 Göttingen, Germany. During this work, he spent six months as visiting researcher at the Clinical Augmented Intelligence Group, Department of Medicine, Massachusetts General Hospital, Boston, MA, USA (e-mail: jonas.huegel@med.uni-goettingen.de).}}

\maketitle

\begin{abstract}
Finding similar patients is a common objective in precision medicine, facilitating treatment outcome assessment and clinical decision support.
Choosing widely-available patient features and appropriate mathematical methods for similarity calculations is crucial.
International Statistical Classification of Diseases and Related Health Problems (ICD) codes are used worldwide to encode diseases and are available for nearly all patients. 
Aggregated as sets consisting of primary and secondary diagnoses they can display a degree of comorbidity and reveal comorbidity patterns. 
It is possible to compute the similarity of patients based on their ICD codes by using semantic similarity algorithms. 
These algorithms have been traditionally evaluated using a single-term expert rated data set.
\\
However, real-word patient data often display varying degrees of documented comorbidities that might impair algorithm performance.
To account for this, we present a scale term that considers documented comorbidity-variance.
In this work, we compared the performance of 80 combinations of established algorithms in terms of semantic similarity based on ICD-code sets.
The sets have been extracted from patients with a C25.X (pancreatic cancer) primary diagnosis and provide a variety of different combinations of ICD-codes.
Using our scale term we yielded the best results with a combination of level-based information content, Leacock \& Chodorow concept similarity and bipartite graph matching for the set similarities reaching a correlation of 0.75 with our expert's ground truth. 
Our results highlight the importance of accounting for comorbidity variance while demonstrating how well current semantic similarity algorithms perform.
\end{abstract}


\section{Introduction}
\label{sec:introduction}
\begin{figure}
    \centering
    \includegraphics[scale=0.5]{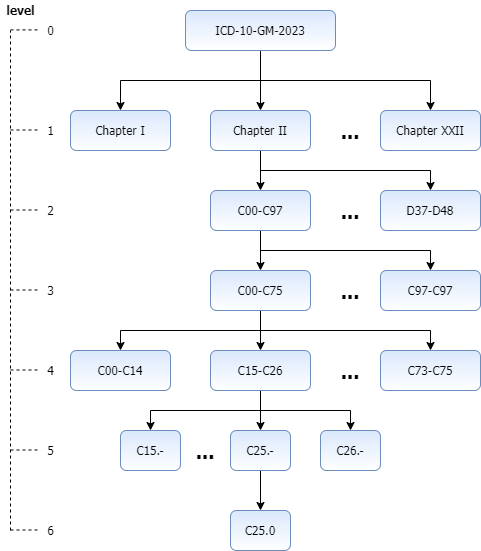}
    \caption{ICD-10-GM 2023 taxonomy with C25.0 ("Malignant neoplasm of head of pancreas") as an example node. The figure depicts the path from root to C25.0 node through the taxonomy tree while showing the node's ancestors. The levels of the respective nodes are shown on the left.}
    \label{fig:icd-fugre}
\end{figure}

A common objective in precision medicine is finding similar patients.
This allows clinicians to collect information on possible treatment outcomes and can aid in other forms of clinical decision support.
Measuring patient similarities and the development and improvement of such measures is an active field of research \cite{Parimbelli.2018, Sharafoddini.2017, McDermott.2021, sathiya_review_2019, hier_evaluation_2020}.
Parimbelli et al. \cite{Parimbelli.2018} and Sharafoddini et al. \cite{Sharafoddini.2017} performed systematic reviews and identified a broad range of data and used algorithms for different areas in medicine.
Both classified the input data into multiple different input classes, which include but are not limited to molecular \cite{andersen_genomic_2012,ross-adams_integration_2015,serra_mvda_2015,kim_integrative_2015}, clinical \cite{jia_patient-similarity-based_2020, hier_evaluation_2020}, and integrated (combined) data types \cite{ow_big_2016,sadanandam_colorectal_2013,hall_predicting_2015}. \\
Using widely available and established patient-features for such calculations is an important step to make research reproducible and applicable in the clinical context \cite{McDermott.2021}.
Besides this, it is just as important to use standardized and easily accessible features to make results comparable and interpretable among clinicians and researchers. 
One feature that meets those requirements is the \gls{icd} taxonomy \cite{who-icd_2023, world_health_organization_icd-10_2004, Harrison_2021}.
It is used world wide to encode diseases and conditions from patients and is available in most hospital data sets.
Many countries use the ICD-encoding for reimbursement \cite{eastern_ohio_health_information_management_association_last_icd-10_2016} which makes its use mandatory for almost all clinical settings in these countries.
Using a medical ontology such as the ICD-taxonomy as a patient-feature for patient similarities has the advantage that the embedded concepts offer a high granularity as well as a standardized set of terms and a standardized structure.
From a computer science point of view the \gls{icd} ontology can be seen as graph or a tree with different nodes that are connected with unweighted edges.
A patient is represented by the set of their \gls{icd} codes.
\autoref{fig:icd-fugre} visualizes this description. 
This allows to apply distance measures from graph theory to calculate a similarity score for a pair of nodes based on their relative position in the ontology.\\
Patients often suffer from comorbidities and are therefore assigned more than one ICD-diagnosis code.
So further combining the pairwise concept-similarities to calculate the similarity of patients' ICD-code sets is a promising approach to yield a more holistic similarity score \cite{li_identification_2015,sarfati_impact_2016}. 

Girardi et al. \cite{girardi_using_2016} used ICD-set similarities to cluster patients with four different main diagnoses.
They showed that their proposed algorithm outperforms algorithms that do not take non-linearities and the hierarchical structure of the ICD-10 taxonomy into account.

Jia et al. \cite{jia_patient-similarity-based_2020} constructed a patient vector based on symptom similarity using one hot encoding to represent whether a certain symptom is present or not.
Additionally, they included lab test similarity, which they calculated through numerical analysis of continuous features and diagnoses similarity based on ICD-set semantic similarity for a diagnostic prediction task.
They used the semantic similarity algorithms that performed best in their previous study \cite{jia_using_2019}.
In their study they analyzed different combinations of \gls{ic}, \gls{cs} and \gls{setsim} algorithms regarding their performance in a classification task for hospital length of stay of nephrology patients.

Wang et al. \cite{wang_study_2021} also used ICD-code-set similarity as a feature in a patient vector to train a semi-supervised k-Nearest-Neighbors (kNN) model for disease prediction.
Their labeled set consisted of 30 randomly chosen patients.
A domain expert scored these patients regarding their clinical similarity.
To calculate ICD-code-set similarity the authors first calculated the \gls{ic} of the nearest common ancestor (NCA) of all concept-pairs from the respective sets.
They then calculated the mean NCA-IC from across the two sets to get a single value that represents the similarity of the two sets.\\
It is to mention, that each of the previously presented works used different selection criteria for the choice of the semantic similarity algorithms and different benchmark-strategies for the performance evaluation. This suggests a lack of research to help identify fitting algorithms for specific use cases in a standardized way.\\
Girardi et al. \cite{girardi_using_2016} did not further elaborate on their choice of the Jaccard and the Haase-Li algorithms to compare their own algorithm and not any other established set similarity algorithm.
Their performance evaluation consisted of visualizing the patient clusters based on the four main-diagnoses. 
There was no ground truth to evaluate how similar the patients are based on their ICD-code sets besides how well they generated the four rather coarse clusters of patients. \\
Jia et al. \cite{jia_patient-similarity-based_2020} did use the algorithm that performed best in their previous survey \cite{jia_using_2019}. 
However, the tested algorithms in their survey \cite{jia_using_2019} have been evaluated using a classification task with four rather unrelated classes. 
The discriminatory power of the evaluated algorithms in the context of their work is therefor to be further tested in other clinical contexts. \\
Wang et al. \cite{wang_study_2021} did discuss their choice of an IC-based algorithm.
They refer to previous works that tested the performance of concept-similarity algorithms such as Sánchez et al. \cite{sanchez_semantic_2011}.
The expert-labeled data used in Wang et al.'s kNN-model represents a valuable gold standard. 
However, they only labeled a small fraction (30 patients) of the $\sim$1000 patients for their semi-supervised-learning approach.
Although they achieved good results for their specific patient similarity application, it can be argued that ground truth generation was again a bottle-neck for this research.\\
Lambert et al. \cite{LAMBERT_2023} worked on a way to improve patient clustering by incorporating structured label relationships in similarity measures.
They extended the cosine-similarity algorithm by applying inverse document frequency and semantic relationships (Wu-Palmer \cite{wu_verbs_1994} \& Lin \cite{Lin_1998}) as weights.
This shows that choosing as well as evaluating semantic similarity algorithms in the biomedical domain remains a challenging task due to the complex nature of ground truth generation for clinical data. 
It also highlights the importance to incorporate semantic information into patient similarity applications.

\subsection{Semantic Similarity Calculations}

Medical knowledge is commonly organized and represented through ontological frameworks, such as SNOMED CT, the Gene Ontology \cite{ashburner_gene_2000}, or the ICD coding system \cite{world_health_organization_icd-10_2004,who-icd_2023,Harrison_2021}.
From an information theoretical point of view, these taxonomies are tree-structures with a root as well as parent, child and leaf nodes. A tree has a certain depth and the deeper a node is situated inside the tree, the more specific it is in this context. \\
Looking at the ICD-10(-GM) taxonomy we can use the code C25.0 (Malignant neoplasm of head of pancreas) as an example to further elaborate the structure. 
The node is a leaf, which means it has no children. 
The path from the ICD-10-GM root to the node is: ICD-10-GM root $\rightarrow$ “Chapter II Neoplasms (C00-D48)” $\rightarrow$  “Malignant neoplasms (C00-C97)" $\rightarrow$ “Malignant neoplasms at specified sites, established or suspected to be primary, excluding lymphoid, haematopoietic and allied tissues (C00-C75)" $\rightarrow$ “Malignant neoplasms of digestive organs (C15-C26)” $\rightarrow$ “Malignant neoplasm of pancreas (C25.-)” $\rightarrow$ "Malignant neoplasm of head of pancreas (C25.0)". \\
This example demonstrates the concept of depth-related specificity in hierarchical taxonomies and is visualized in \autoref{fig:icd-fugre}. 
These structures represent the relatedness of the underlying concepts and make it possible to perform semantic similarity calculations.
There are multiple approaches to calculate semantic similarity between two concepts in an ontology. 
Those approaches can be classified broadly into two classes: the first one being ontology- or path-based algorithms that take the ontological structure into account. 
A trivial example for such an algorithm would be the “shortest path” approach. 
Two concepts are more similar/less distant when the path to get from concept A to concept B is shorter when compared to concepts with longer distances. 
The second class being information-content (IC) based approaches. These approaches take into account how “specific” a certain concept is. 
The general idea behind the use of a concept’s IC is that a more specific (as in for example deeper in the taxonomy tree) a concept is, the more information it provides. 
One can then take this specificity into consideration when calculating the similarity of two concepts in an ontology. 
An early example for such an approach would be Resnik’s similarity measure\cite{resnik}. 

Research has been conducted to evaluate which of these approaches yields better results in the biomedical context and it has been shown that the IC-based approaches outperform ontology-based methods \cite{althobaiti_comparison_2017,sanchez_semantic_2011,boriah_similarity_2008, garla_semantic_2012}.
Sánchez and Batet presented IC-based redefinitions of established ontology-based measures that yielded better results than their “original” counterparts \cite{sanchez_semantic_2011}. 
IC-based concept-similarity (CS) measures require – as the name implies – the calculation of the IC. 
This means that researchers seeking concept similarities might have to decide for a tuple (IC, CS) for their specific application.
A straight-forward approach to calculate the specificity and IC of a concept would be to use its depth or level in the taxonomy tree.
Other more sophisticated IC-measures take the underlying structure of the given ontology into consideration \cite{batet_ontology-based_2011,mazandu_information_2014}.

\subsection{Semantic similarity in the biomedical domain}
Applying these measures in the biomedical context leads to the observation that most patients have more than one diagnosis or that a certain gene of interest is annotated with more than one GO term \cite{jia_using_2019,mazandu_gene_2016}. 
This makes it important to include another type of algorithm to ensure a holistic approach to the similarity calculation. 
Set-similarity (SetSim) algorithms allow for such a contextualization of concept-level similarities \cite{jia_using_2019,girardi_using_2016}.
Established and rather “trivial” algorithms are for example Dice or Jaccard similarities, which in one way or another offer a  “hard” similarity coefficient that takes into account the intersection of two concept-sets. 
Those algorithms are computationally efficient since they do not require any further calculations (besides comparing which concepts are present in the sets). 
Hard in this context means that the intersection is calculated in a way where two concepts either exist in both sets or they do not. 
This completely undermines the fact that two sets might consist of concepts that are really similar but not exactly the same. \\
Regarding taxonomy-based similarity calculations, more sophisticated approaches that take the taxonomic structure into account might be more fitting \cite{jia_using_2019,girardi_using_2016}, depending on the application.
Other “softer” approaches can be some variation of averaging over the CSs of the sets\cite{jia_using_2019,mazandu_information_2014} or weighted bipartite graph matching where the Hungarian/Kuhn-Munkres-algorithm can be used to either maximize (for similarity measures) or minimize (for distance measures) the intended measure \cite{jia_using_2019}. 
This means that to compare i.e. a set of ICD-codes of two patients, one might be in need of a triplet (IC, CS, SetSim) of algorithms to calculate a similarity score. 
The actual algorithms in such a triplet strongly depend on the use case and the biomedical niche where the measures are supposed to be applied \cite{jia_using_2019,mazandu_information_2014,mazandu_gene_2016, kulmanov_semantic_2021,girardi_using_2016,althobaiti_comparison_2017}. 
This indicates that it is important to perform benchmarks in each subdomain to find out which of the current algorithms offer the greatest use for a certain application.
A problem researchers face in this context is the lack of ground truths for the evaluation of such calculations.
Pairwise comparisons of patients based on their medical history are time consuming.\\
One often used public data set for such benchmarks is Pedersen et al.'s set of 30 medical concept pairs that have been scored by three physicians and nine medical index experts \cite{pedersen_measures_2005, abdelrahman_survey_2015, al-mubaid_measuring_2009, nguyen_mubaid_2006, m_b_abdelrahman_evaluating_2018, mohd_survey_2013}. 
This set offers similarity scores ranging from 0.0 to 4.0 for the medical term pairs. 
Although useful for single-term comparisons, Pedersen et al.'s test bed does not offer scores for sets of medical terms.
This lack of "gold standards" makes it even more challenging to compare set-similarity algorithms across different biomedical domains.

\subsection{Contribution}
In this paper we analyze how well the state-of-the-art semantic similarity algorithms perform on sets of ICD-10 codes with a special focus on their ability to account for the degree of documented comorbidity in patients. \\
The ICD-code sets are derived from patients with a "Malignant neoplasm of pancreas" (C25.X) main diagnosis. 
Initially, we evaluated 40 algorithm-combinations consisting of different information content, concept-similarity as well as set similarity measures. 
We then compared those results with similarity scores determined by an oncologist.\\
Our results confirm the need for consideration of documented comorbidity-degrees reflected by the size of the ICD-code sets.
To account for them, we present a scale term which improved algorithm performance and yielded the best results among the total of 80 algorithm-combinations.

\section{Methods}
\label{sec:methods}

\subsection{Data Acquisition}
\label{subsec:data_acq}
The patient data we used for our calculations have been collected in the context of the \gls{mtb} at the \gls{umg}. 
The data contains the main diagnoses the patient was treated for at the \gls{umg} \gls{mtb} in the period from 2019 to 2022, and additional known diagnoses from the patient in the \gls{ehr}.
We decided to include all patients with a C25.X primary-diagnosis and gathered all unique secondary diagnoses of these patients that were available in the \glspl{ehr}. 
All codes in these patient-diagnoses sets have been weighted equally in our calculations. 
As a result, our data set consists of 29 patients with a mean of $\approx35$ \gls{icd}-codes (median = 31) and a standard deviation of $\approx24$ codes.
The smallest sets consisted of only one \gls{icd} code whereas the largest set contained 94 diagnoses codes. 
This suggests a high degree in documented comorbidity-variance between the patients and highlights the importance to account for this.
We did not exclude any \gls{icd}-chapter but instead used all available diagnoses codes.\\
In addition, we assigned pseudonyms to each patient(-set) linking back to the original patient-id. 
The ethic committee's vote for the \gls{umg} internal \gls{mtb}s allows retrospective analysis of the patient data. 
Additionally, all \gls{mtb} patients with a C25.X diagnosis are also enrolled in the Molecular Pancreatic Cancer Program of the \gls{umg} (MolPAC), which has an additional ethic committee vote also allowing retrospectives analysis.\\
To be able to compare the performance of the algorithms we needed some form of ground truth. 
We generated a ground truth similarity-matrix by handing a list consisting of patient pseudonyms and their corresponding sets of \gls{icd}-codes to a pancreatic-cancer expert. 
Based on this list of codes (and only based on them) they generated a matrix (\autoref{fig:expert_matrix}), which we proceeded to compare to the matrices generated by the algorithms. 
The similarity scores in this matrix ranged from 0 (no similarity) to 10 (identical).

\subsection{Algorithm Aggregation}
\label{subsec:algoAggregation}
To be able to compare available algorithms regarding their performance, it was important to get an overview of the state-of-the-art semantic similarity algorithms. 
We searched for semantic similarity algorithms that have been used in the biomedical context.
An overview of all algorithms used in this study can be seen in \autoref{tab:algorithm_overview}. \\
On the IC level we decided to include the computationally more efficient but less sophisticated level-based algorithm (\#1). 
With this approach, the IC of a concept is equal to its depth in the taxonomy-tree (compare with \autoref{fig:icd-fugre}).
We also included Sánchez et al.’s IC-measure (\#2)  \cite{sanchez_semantic_2011, sanchez_ontology-based_2011} that takes the taxonomic structure into account. 
Choosing these two algorithms allows for a good comparison between efficiency and computational complexity. 
\\
For CS, we decided to include the algorithms from Nguyen \& Al-Mubaid (\#3) \cite{nguyen_mubaid_2006} \& Li (\#7) \cite{yuhua_li_approach_2003} since they clearly outperformed others in previous works \cite{althobaiti_comparison_2017,kulmanov_semantic_2021,m_b_abdelrahman_evaluating_2018}. 
Additionally, we included Sánchez et al.'s redefinitions \cite{sanchez_semantic_2011} of the CS algorithms by Leacock \& Chodorow (\#5) \cite{Leacock1998CombiningLC} \& Wu-Palmer (\#8) \cite{wu_verbs_1994} due to their performance advantages regarding similarity scoring \cite{mohd_survey_2013}.  
As stated by Sánchez et al. \cite{sanchez_semantic_2011}, their redefined Wu-Palmer measure is similar to Lin's similarity measure \cite{lin_principle-based_1993}. 
We also decided to include a simplified version of the Wu-Palmer measure (\#6) which showed good results while being more computationally efficient \cite{jia_using_2019}. 
For completeness we also included one edge/path-based algorithm (\#4) as a comparison to the ontology-based algorithms.\\
For the SetSim algorithms we included mean CS (\#9), weighted bipartite graph matching (\#10) and the hierarchical distance measure (\#11) \cite{girardi_using_2016}, due to their performance in previous works \cite{jia_using_2019,girardi_using_2016}.\\
An important point to consider when using weighted bipartite graph matching (\#10) is the requirement to adjust for the used CS metric regarding minimizing or maximizing the weight score. 
If a CS algorithm calculates distances one would be needed to use \textit{minimum} matching to match the "closest" concepts. 
If the CS algorithms returns similarity scores one would need to use \textit{maximum} matching to retrieve the highest similarity score for two sets of concepts. \\
The hierarchical distance algorithm (\#11) demands distance scores as input \cite{girardi_using_2016}.
Therefore, it is required to either use native distance metrics or to convert similarity scores to distances.
This can be done by normalizing the results to have a value between 0 and 1, and afterwards subtracting this normalized similarity score from 1 \cite{sidorov_soft_2014}.
Last but not least, we decided to include these more common and rather trivial and therefore less computationally intensive similarity coefficients: overlap (\#12), cosine (\#13), dice (\#14), jaccard (\#15).
Similar to the IC-algorithms choice, this allows for a good comparison between efficiency, computational complexity and the efficacy of a given algorithm.

\subsection{Influence of set size on performance}

Since the degree of documented comorbidity and therefore the size of diagnoses sets can vary heavily we decided to develop a scale term \eqref{eq1} that regulates the influence of the set sizes.
\\
To demonstrate this influence, let us consider the following extreme case using the maximum bipartite matching algorithm: 
Assume that $A$ and $B$ are sets with 100 codes each, so that $|A| = |B| = 100$.
Furthermore, let us assume that every pairwise comparison of a concept $a_i \in A$ with a concept $b_j \in B$ yields a low similarity score of 0.1.
With this, for each pair ($a_i,b_j)$: $CS(a_i,b_j) = 0.1 $ holds, marking each concept pair as highly distinct. 
When applying maximum bipartite matching, one would still get a similarity score of 10, since we needed to match every code of set A with one of set B, $SetSim_{mbm}(A,B) = 10$.

Now, let us compare this result with $SetSim_{mbm}(C,D)$, where $C$ and $D$ are two other sets both consisting of 5 codes each, $c_k \in C$ and $d_l \in D$. \\
Under the assumption that the two smaller sets $C,D$ are being highly similar with all pairwise comparison yielding high scores,  $CS(c_k,d_l) = 0.9$.
That results in a similarity score $SetSim_{mbm}(C,D)= 4.5$ marking these two sets $\sim$50\% less similar than the first two, even if the entries of these sets have significantly higher similarity.\\
Set size similarity per se is of course a factor that has to be included and rewarded in these calculations.
Therefore, we provide a scale term $ST$ that regulates the impact the set size has on the results and as we show later significantly improves the algorithms' performance with our data set.
We define the scale term $ST$ as follows: 
\begin{equation}\label{eq1}
    ST(A,B) = \frac{\text{setSim}(A, B)}{\min{(|A|,|B|)} + \log(1 + \text{abs}(|A|-|B|))}
\end{equation}
where the variables $A$ and $B$ denote the two sets that are being compared and both consist of the ICD codes $a_i \in A$ and $b_i \in B$.
$setSim(A,B)$ is the result of the unscaled similarity calculation for the two sets whereas $\min{(|A|,|B|)}$ returns the set size of the smaller set of the two.\\
The idea is to scale the similarity in a logarithmic way while also accounting for the difference in set sizes. 
Since the smallest set size presents the upper bound for the similarity score it is also important to take this value into account by using it as reference/normalization factor. 
The scale term still provides a higher similarity for larger sets with similar entries, than for smaller sets with the same number of similar entries.
This holds significance as it renders them comparatively more similar in contrast to smaller sets.
One could argue that bigger sets have a higher possibility of being similar. 
At the same time it is just as important to account for the fact that bigger sets by default might end up with higher scores depending on the applied  algorithm. 
Our term is supposed to account for exactly this "vulnerability" while preserving a certain perk for similarities in the set size.

\begin{table*}[h]
\caption{Overview table showing the used Information Content (IC), Concept Similarity (CS) and Set Similarity (SetSim) algorithms. For each algorithm we provide the type, the number used for reference in our study, a short definition and their literature reference. 
The definitions are kept short in this table, a more detailed explanation can be found in the corresponding references.
Moreover, we omitted the definitions for \#10 and \#11 due to their complexity and refer the reader to the original paper.
The design of this table is inspired by Jia et al.'s overview \cite{jia_using_2019}.}
\label{tab:algorithm_overview}
\bgroup
\def\arraystretch{2}
\begin{adjustbox}{width=\textwidth}
\begin{tabular*}{\textwidth}{c|c|c|c|l}
Type of Algorithm        & \# & Algorithm                & Definition & Reference \\ \hline
Information Content (IC) & 1    & level-based              & $level(c)$            &     --      \\
                         & 2    & Sánchez et al.           &  $-\log{\frac{\frac{\#leavesSubtree}{\#subsumers}+1}{\#allLeaves}} $        & \cite{sanchez_ontology-based_2011}\\ \hline
Concept Similarity (CS)  & 3    & Nguyen \& Al-Mubaid      & $\log((shortestPath(c_1,c_2) - 1) * (depth_{max} - depth_{lca}) + 1)$           & \cite{al-mubaid_measuring_2009} \\
                         & 4    & path based               &  $\frac{depth_{max}}{depth_1 + depth_2} $         & -- \\
                         & 5    & Leacock \& Chodorow      &  $-log( \frac{ic_1+ic_2-2 * ic_{lca+}1}{2*ic_{max}})$          & \cite{Leacock1998CombiningLC} \\ 
                         & 6    & simple Wu-Palmer         &  $1 - \frac{depth_{max} - ic_{lca}}{depth_{max}}$          & \cite{jia_using_2019} \\
                         & 7    & Li                       & $\frac{\exp{(0.2 * (ic_1 +ic_2 - 2*ic_{lca}))} * (\exp{(0.6 * ic_{lca})} - \exp{(-0.6 * ic_{lca})})}{\exp{(0.6 * ic_{lca})} + \exp{(-0.6 * ic_{lca}})}$           & \cite{yuhua_li_approach_2003} \\
                         & 8    & Wu-Palmer                & $\frac{2 * ic{lca}}{ic_1 + ic_2}$           & \cite{wu_verbs_1994}   \\ \hline
Set Similarity (SetSim)  & 9    & mean CS                  &  $\frac{(\sum_{c1_i \in set_1}\sum_{c2_i \in set_2} CS(c1_i,c2_i)) * 0.5}{|set_1| + |set_2|}$           & \cite{jia_using_2019} \\
                         & 10   & weighted Bipartite Graph Matching   &   see reference        & \cite{jia_using_2019}  \\
                         & 11   & Hierarchical Distance & see reference           & \cite{girardi_using_2016} \\
                         & 12   & Overlap                  & $\frac{|set_1 \cap set_2|}{min(|set_1|, |set_2|)}$           &     --      \\
                         & 13   & Cosine                     & $\frac{|set_1 \cap set_2|}{\sqrt{|set_1| * |set_2|}}$           &      --     \\
                         & 14   & Dice                  & $\frac{2* |set_1 \cap set_2|}{|set_1| + |set_2|}$           &       --    \\
                         & 15   & Jaccard                   &   $\frac{|set_1 \cap set_2|}{|set_1 \cup set_2|}$         &     --     
\end{tabular*}
\end{adjustbox}
\egroup

\end{table*}

\subsection{Study Design}
We created 4-tuples $(IC,\ CS,\ SetSim,\ scaled/unscaled)$ of all possible algorithm-combinations and used them to calculate the set similarities for all 29 patient-sets.
Each combination has been calculated once with and once without our scaling term.
We then calculated the pearson correlation coefficient of the calculated matrices and the expert matrix.
For the similarity calculations we used Python 3.9.6 with taxodist \cite{taxodist_git}, an open source library we developed to perform similarity calculations.
We also uploaded a snapshot \cite{taxodist_snapshot} of the used taxodist version and our analysis code to the Göttingen Research Online Data repository, so that our results can be recreated and tested for validity as well as to enable others to use our library. 
We used pandas 1.3.0 \cite{reback2020pandas} and sci-kit learn 1.1.1 \cite{scikit-learn} for the statistical analysis as well as seaborn 0.11.2 \cite{Waskom2021} for the visualization of the results. \\
It is important to mention that each hospital (group) has their own specific set of available data. 
The sensitivity of medical complicates sharing and publishing it.
These circumstances also makes a comparison of the applied algorithms to detect similarity more challenging \cite{McDermott.2021}.
The limited amount of publicly available data sets might allow to benchmark the algorithms using them, but the risk of overfitting and bias towards the structure and data  of these public data sets is considerably high \cite{McDermott.2021}.
The same applies to our local data that was used to benchmark the algorithms.

\section{Results}
\label{sec:results}

\begin{figure*}
    \centering
    \includegraphics[scale=0.46]{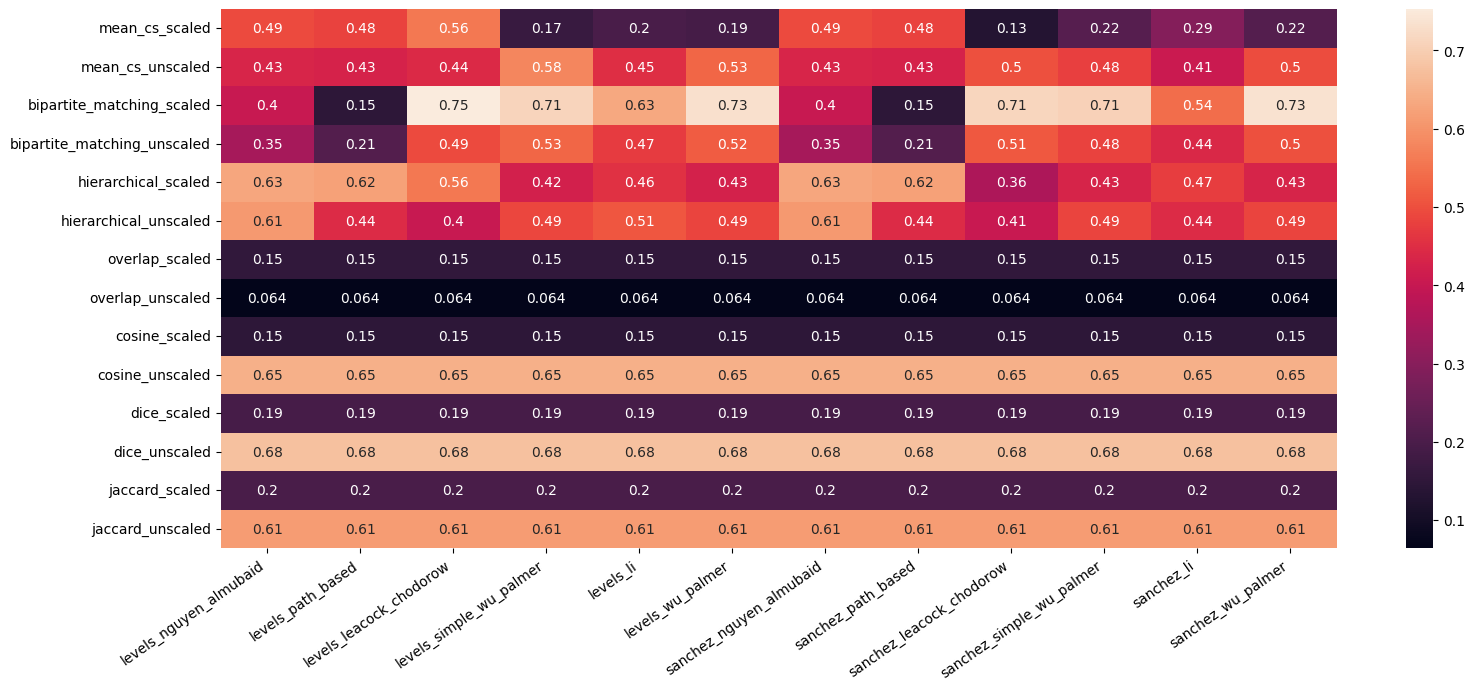}
    \caption{Heatmap displaying the correlation values of all possible combinations of algorithms. The y-axis shows the set-level algorithm and whether the calculation has been scaled to the set size or not. The x-axis shows the combination of IC and CS algorithms.\\\\}
    \label{fig:overview_heatmap}
\end{figure*}

A complete heatmap of all 80 combinations and the corresponding correlations with the expert's ground truth can be seen in \autoref{fig:overview_heatmap}. We provide a detailed enumeration of the correlation, as well as the similarity scores for each algorithm-combination in the Appendix.

\begin{figure*}[]
\centering
    \begin{subfigure}[t]{0.43\textwidth}
        \includegraphics[width=\textwidth]{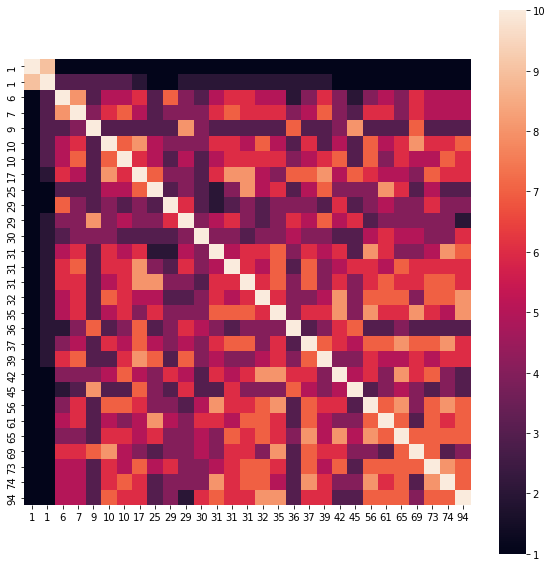}
        \caption{Expert-matrix used as ground truth for all correlation calculations.}
        \label{fig:expert_matrix}
    \end{subfigure}
    \hfill
    \begin{subfigure}[t]{0.43\textwidth}
        \includegraphics[width=\textwidth]{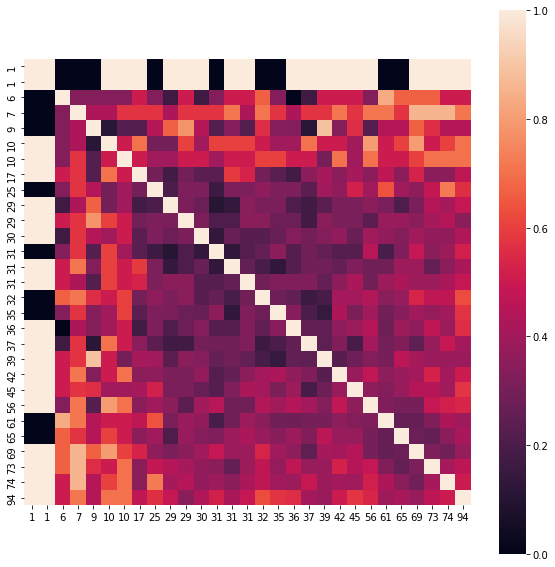}
        \caption{Similarity matrix for worst performing algorithm \textit{(overlap,unscaled)} (correlation 0.064).} 
        \label{fig:worst}
    \end{subfigure}
    \hfill
    \begin{subfigure}[t]{0.43\textwidth}
        \includegraphics[width=\textwidth]{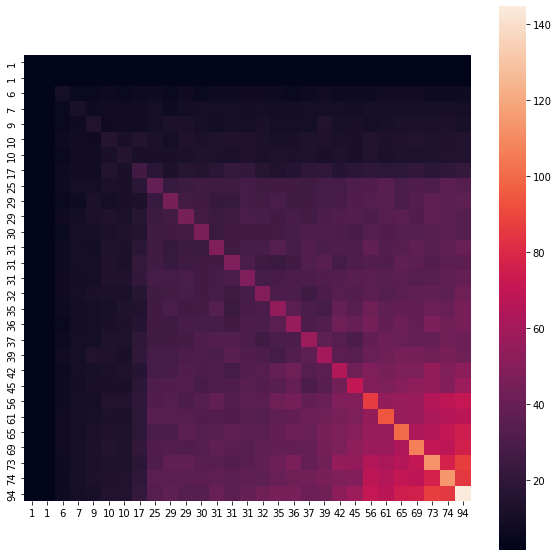}
        \caption{Similarity matrix for \textit{(level-based, Leacock \& Chodorow, bipartite matching, unscaled)} (correlation 0.49). This algorithm combination performs subpar without scaling as can be seen in \autoref{fig:best_scaled}.}
        \label{fig:best_unscaled}
    \end{subfigure}
    \hfill
    \begin{subfigure}[t]{0.43\textwidth}
        \includegraphics[width=\textwidth]{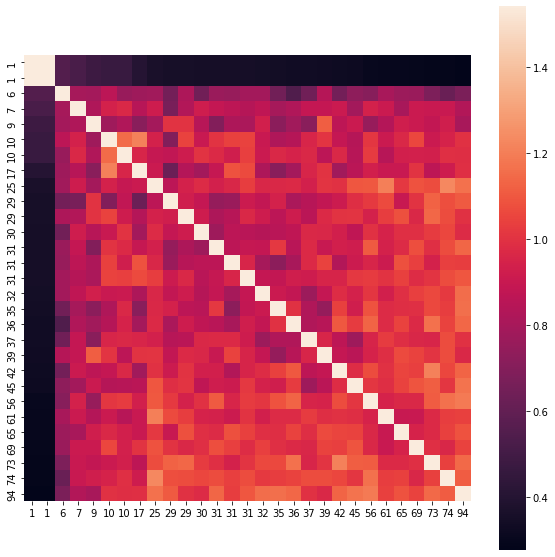}
        \caption{Similarity matrix for best performing combination after scaling \textit{(level-based, Leacock \& Chodorow, bipartite matching, scaled)} (correlation 0.75).}
        \label{fig:best_scaled}
    \end{subfigure}
    \caption{Patient-similarity-heatmaps for ground truth \& different algorithms showing the scores for the pairwise patient similarity. The x and y axes are showing the corresponding size of their ICD-code sets.}
    
\label{fig:result_heatmaps}
\end{figure*}

The best results were achieved with the 4-tuples \textit{(level-based, Leacock \& Chodorow, bipartite matching, scaled)} achieving a Pearson correlation of 0.75 (\autoref{fig:best_scaled}) and \textit{(level-based, Wu-Palmer, bipartite matching, scaled)} achieving 0.73.
It is to be noted that the "more trivial" algorithms  (Cosine, Dice, Jaccard) all performed well with respective correlation values of 0.65, 0.68 and 0.61. 
They outperformed most of the computationally more intense algorithm-combinations.
\\ \\
When comparing the different algorithmic levels of the 4-tuples, our results suggest that they have varying impacts on the correlation score. 
For the IC-algorithms, it is to say that the two algorithms did not yield significantly varying results. 
With our data and the underlying taxonomy it seems that the computationally more efficient and less complex level-based IC even outperforms Sánchez IC measure on average ($\approx9\%$ better).\\
In terms of CS-algorithms we found the redefined Wu-Palmer and Leacock \& Chodorow to perform best. However, our results suggest that for every CS-algorithm there is one combination that allows for a relatively high correlation ($>$0.5). \\
Regarding the set-similarity algorithms our results show that the bipartite matching algorithm performed best but that the more trivial algorithms (Cosine, Dice, Jaccard) do not fare badly in comparison.
\\
The introduction of our scaling term improved some of the combinations and provided the best overall results while worsening others.
The algorithms that benefited from scaling are those that did not account for set size initially, whereas those with worse performance after scaling already did.
The mean CS (\#9) for example already scales by default since it averages the given CS similarities in the sets. Although we see improvements for combinations with Nguyen-Al-Mubaid, path-based and Leacock \& Chodorow, this algorithm performs best without scaling  (highest correlation = 0.58 for \textit{(level-based, simple Wu-Palmer, mean CS, unscaled)}).
\\
It is to mention that the combination that benefited the most of the scale term was the one that correlated the best with the expert’s rating (\autoref{fig:best_unscaled},\autoref{fig:best_scaled}).
The algorithm with the worst performance was the unscaled overlap algorithm (\autoref{fig:worst}).

\section{Discussion}
\label{sec:discussion}
\subsection{Related work}
To contextualize our work, we compare our approach and results with other semantic similarity benchmarks from the biomedical domain.
Mohd et al. \cite{mohd_survey_2013} compared eight measures including Li, Leacock \& Chodorow and Wu-Palmer.
They compared their performances on the Pedersen data set \cite{pedersen_measures_2005} but only used the physician-ranks without further elaborating on this choice.
The algorithm that performed best in their work was Leacock \& Chodorow's CS, yielding a correlation value of 0.733.
This result differs significantly from our best unscaled Leacock \& Chodorow CS (0.51) result.
Using our scale term we improved the algorithm's performance to a comparable correlation of 0.75. 
Althobaiti \cite{althobaiti_comparison_2017} compared the performance of five CS measures using Pedersen's term-set \cite{pedersen_measures_2005}.
The best performing algorithm here was Nguyen-Al Mubaid's measure.
It outperformed the path-based measure as well as Wu-Palmer's, Lin's and Leacock \& Chodorow's.
Althobaiti and Mohd et al. exclusively compared the performance of CS-algorithms on the concept-level whereas our results offer an overview of IC, CS \& SetSim algorithm combinations in the biomedical context regarding pancreatic cancer patients.
This explains the divergence in the direct comparison with the results of our work.
Rather than offering a "winning" combination, we want to emphasize the importance of contextualization of the applied algorithms.
It is also important to keep in mind, that most CS algorithms achieved good results depending on the other used algorithms and whether or not scaling was applied.
Thus, it is further important to look at works that compared set similarity algorithms in the biomedical context.
Girardi et al. \cite{girardi_using_2016} compared their proposed hierarchical-distance algorithm for set-similarities with the Jaccard and the Haase-Li distances \cite{haase_hutchison_peer_2004}.
They found their algorithm to outperform the other two in the clustering task.
As CS measure they used the path-based algorithm. 
Our results suggest that the use of our scaling-term as well as other CS-algorithms improve the algorithm's performance.
The best results for their algorithm have been achieved using Nguyen-Al Mubaid's CS measure \cite{nguyen_mubaid_2006} in combination with our scale term, yielding a correlation value of 0.63.
This combination also performed well without scaling (0.61).
The CS-algorithm they used in their original evaluation did not perform well in combination with the hierarchical distance algorithm without scaling (0.44).
However, the use of our scale term improved the performance and raised the correlation to 0.62.

Jia et al. \cite{jia_using_2019} compared different IC, CS and SetSim algorithms in the context of a hospital length of stay (HLOS) prediction task based on patients' ICD-10 code sets.
Their findings suggest that the level-based IC algorithm is a good choice for narrow concept backgrounds.
Our results confirm this since our data are taken from a narrow concept background (all main diagnoses in the context Pancreatic Cancer/C25.X).
However their findings considering the use of the bipartite graph matching algorithm do not coincide with our results.
They suggest that the bipartite graph matching algorithm would be better suited when the sizes of two sets are large or unbalanced.
Our data set is highly heterogeneous regarding set size as we explained in \autoref{subsec:data_acq}.
Nevertheless, the bipartite graph matching algorithm performs only mediocre without the application of our scaling term.
However, the use of our scale term shows that accounting for differences in set size improves the algorithm's performance.
It is also to be noted that the unscaled bipartite matching algorithm was outperformed by other SetSim-algorithms like mean CS, e.g. \textit{(level-based, simple Wu-Palmer, mean CS, unscaled)} with 0.58, or the hierarchical-distance algorithm e.g. \textit{(level-based, Nguyen \& Al-Mubaid, hierarchical-distance, unscaled)} with 0.61.
This can be explained by the different tasks the algorithms were used for.
Jia et al. \cite{jia_using_2019} used the algorithms to classify patients in four classes depending on their HLOS and their age whereas we benchmarked the algorithms on the basis of an expert-generated ground truth.
Those two use cases might be too different from each other to be comparable in the context of set-similarity algorithm evaluation.
In addition to those works, Lambert et al. \cite{LAMBERT_2023} developed a weighted cosine-similarity measure that extends the established cosine-similarity algorithm by incorporating semantic similarity measures (Wu-Palmer and Lin) as weights.
Their findings coincide with ours in that the cosine-similarity algorithm is not improved by applying the suggested scaling-approach in the context of general population.
\subsection{Influence of baseline patient similarity on algorithm performance}
In this work, we have evaluated how well state-of-the-art IC, CS and SetSim-algorithms perform on sets of ICD-codes compared to an experts rating. 
Our results suggest that the combination of a level-based IC, Leacock \& Chodorow's CS calculation and bipartite graph-matching on the set-level provides the highest correltation (0.75) with an experts opinion. 
It is also to be noted, that less computationally intense algorithms like Cosine (0.65), Dice (0.68) or Jaccard (0.61) set-similarities result in significant correlations making them interesting options for applications with less computational power/time available.
Possible reasons for the good performance of these algorithms might be the set size as well as the fact that the sets have been derived from already relatively similar patients.
The set size influences the possibility of a concept occurring in a set. 
The more codes a set holds on average, the higher the possibility that there might be a match with a concept from another set.
The high baseline similarity of the patients we chose stems from the fact that patients with similar main diagnoses (in this case C25.X) will end up with similar symptoms leading to similar ICD-10 codes.
This again increases the possibility of concept-matches between sets of codes. \\
The described assumption that clinically similar patients will have similar ICD-codes is a premise for this kind of research.
At the same time it is also a challenge for our work since it is important to find a balance regarding the patients' baseline similarity.
The challenge is to choose a cohort that provides a baseline similarity that allows for a good test of an algorithm's discriminatory power.
Using codes from patients that are highly distinct - e.g. by comparing patients from differing medical departments - might impede the analysis of an algorithm's discriminatory power since the similarity-scoring-task is not granular enough.
However, if the patients already have a high baseline similarity - as could be argued for our data - calculating similarity purely based on their ICD-codes becomes a more difficult task.
This highlights the importance of patient-feature vectors that incorporate different clinical attributes to represent a patient's health status.\\
Another aspect that influenced our cohort design is to make ground truth generation feasible.
Pairwise comparison of patients in and of itself is a time-consuming task since one is in need of $ \frac{n^2}{2}$ comparisons, where n is the number of patients.
Basing these comparison on sets with up to 90 codes makes this even more tedious and complex.
Comparing patients from different medical domains would require experts from different fields working together to generate similarity scores or an expert proficient in those fields.
Hence our choice to narrow our scope down to pancreatic cancer patients to allow our expert to make the comparison feasible.
\newline
Our goal was to find an algorithm (-combination) that would perform best given our data as an exemplary "raw" set of patient ICD-codes. 
We found that accounting for set size is an important step for most algorithms and that different algorithm combinations work better then others.
We also demonstrated the usefulness of scaling to the degree of documented comorbidity to increase algorithm performance.
We decided not to include mean values or other measures that describe an algorithms "average" performance. 
This is due to the fact that those descriptive indicators are not meaningful in this context. 
Our findings support, as reported by others (compare \cite{LAMBERT_2023, jia_using_2019}),  that the performance of any algorithm(-combination) depends on the underlying taxonomy as well as baseline similarity of the respective cohort and their set sizes.
Since set size represents the degree of comorbidity and the algorithm's performance is influenced by this metric, it is important to consider this factor during algorithm selection. 

\subsection{Defining meaningful concepts}
Since there are different available ICD-10 versions (GM, CM, ..), it is important to keep track of the respective coding system that was used for symptom and disease encoding.
This recommendation holds for all ontologies and is an important factor to allow for comparability and reproducibility.
An example as to why this is important can be the C25.0 concept ("Malignant neoplasm of head of pancreas").
In the ICD-10-GM this node would have a depth of six, whereas in the ICD-10-CM it would have a depth of four.
This circumstance would require researchers that perform semantic similarity calculation on cohorts from countries using differing coding systems to harmonize the data before comparing the cohorts.
In our work we used "raw" ICD-10-GM code sets. "Raw" in this context means that we included all documented codes for each patient over the time of their treatment.
It might be interesting to consider excluding certain ICD-codes and maybe even whole ICD-chapters or - more generally speaking - taxonomic concepts and branches from the calculations.
This might help to limit the scope to symptoms/diseases that are relevant for the use case.
To our best knowledge, this kind of "taxonomy pruning" has not been done and compared to the use of "raw" ICD-code sets.
We decided not to apply this technique to our data set.
Our goal was to benchmark the different algorithm-combinations on a holistic data set, since we wanted our expert to also rate the patients based on the complete list of their ICD-codes. 
Another approach to make sure only relevant concepts are considered for similarity calculations would be to define a set of relevant concepts beforehand.
Such a list would have to be generated by experts from the medical field and be supported by clinical research outcomes to ensure, that a certain symptom or diagnosis is actually relevant for a disease or use case.
\\
Not working with raw ICD-sets might therefor avoid "dilution" of similarity scores by incorporating too broad or too often occurring terms.
Examples for such terms from our data set are Z11 ("Specific procedures for testing for infectious and parasitic diseases") and U99.0 ("Special procedures for testing for SARS-CoV-2") which appeared in 26 and 23 sets respectively across our 29 sets. 
Additionally, this highlights the importance of choosing patients from similar time frames.
When using patients sets from a larger time frame harmonization steps might be required.
It is safe to assume that these codes have been added during the COVID-19 pandemic and that it might make sense to exclude such procedure codes depending on the context of the research, since they might not offer an information gain due to their routinely nature during the COVID-19 pandemic.
This highlights the importance to include domain experts to define which concepts provide useful insights and which should be excluded from semantic similarity calculations.

\subsection{Set size and degree of comorbidity}
In our work we equated a patient's ICD-code set size with the degree of documented comorbidity.
The formulation "documented" is important in this context, since there are several factors that influence how many ICD-codes are documented for a certain patient. 
The set of codes is merely a representation of how well documented the health status of a patient is and should not be confused with actual comorbidity since we do not know of a certain condition is still undocumented.
It is important to consider which factors influence the set size and how much explanatory power the amount of documented ICD-codes have.
A patient that has been longer under treatment and had to attend a clinician more often will end up with more codes.
The number of ICD-codes in a patient set might be influenced by the distances from the living place of a patient.
Patients living directly in Göttingen might be treated more often at the UMG compared to patients who live in the surrounding cities. 
They might only be seen for urgent and more complicated cases at the UMG, since the UMG is a maximum care provider. 
Additionally, each country has different regularizations ICD-code-documentation.
In our case, since we are using claims data, we might miss ICD codes which were not relevant for accounting.
Moreover, ICD-coding is a clinicians responsibility which makes it subject to differences in coding habits or personal preferences between different clinicians.
All these factors are import to realistically assess the reliability of the ICD-code sets for patient similarity calculations.
\\
\\
In the future, it could be beneficial to include a broader range of diseases and patients to increase the generalizability of our findings.
One possibility would be to include patients from the same department but with different main-diagnoses.
This would granularly increase the scope while allowing for a more holistic approach.
Overall, this research contributes to the advancement of precision medicine by providing insights into the performance of semantic similarity algorithms for identifying similar patients based on ICD codes. 
The findings underscore the necessity of considering comorbidity variance in real-world patient data, enhancing the accuracy and applicability of similarity calculations. 
These outcomes have implications for improving patient stratification, personalized treatments, and disease understanding, ultimately leading to better healthcare outcomes in precision medicine.

\section{Conclusion}
\label{sec:conclusion}
In conclusion, our work addresses the crucial objective of finding similar patients in precision medicine. 
By utilizing widely available patient features and fitting mathematical methods for similarity calculations, this research focuses on ICD-codes as a means to encode diseases and reveal comorbidity patterns. 
We introduced a scale term that considers documented comorbidity-variance, recognizing the varying degrees of comorbidities in real-world patient data that can impact algorithm performance.
We compared the performance of 80 combinations of established algorithms, specifically semantic similarity algorithms, based on ICD-code sets extracted from patients with pancreatic cancer (C25.X) as their primary diagnosis. 
Through the utilization of the scale term, the study identifies the best-performing combination of level-based information content, Leacock \& Chodorow concept similarity, and bipartite graph matching for set similarities. 
The results demonstrate a significant correlation of 0.75 with the experts' ground truth, emphasizing the importance of accounting for comorbidity variance and highlighting the efficacy of current semantic similarity algorithms.
\bibliographystyle{IEEEtran}
\bibliography{references}
\end{document}